\documentclass{article}

\usepackage[preprint]{neurips_2026}

\usepackage[utf8]{inputenc} 
\usepackage[T1]{fontenc}    
\usepackage{hyperref}       
\usepackage{url}            
\usepackage{booktabs}       
\usepackage{amsfonts}       
\usepackage{nicefrac}       
\usepackage{microtype}      
\usepackage{xcolor}         

\usepackage{floatrow}
\newfloatcommand{capbtabbox}{table}[][\FBwidth]

\usepackage{graphicx}
\usepackage{subfigure}
\usepackage{wrapfig}
\usepackage{amsfonts}
\usepackage{amsmath}
\usepackage{algorithm}
\usepackage{algorithmic}
\usepackage{color}
\usepackage{booktabs}
\usepackage{multirow}
\usepackage[normalem]{ulem}
\useunder{\uline}{\ul}{}
\usepackage{array}
\usepackage{colortbl}
\usepackage{xcolor}
\usepackage{caption}
\usepackage{wrapfig}
\usepackage{subcaption}
\usepackage{array,booktabs,makecell}
\title{Head-Aware Key-Value Compression for Efficient Autoregressive Image Generation}

\author{%
  Guotao Liang\textsuperscript{1,2}, Baoquan Zhang\textsuperscript{1}\thanks{Corresponding Authors}, Zhiyuan Wen\textsuperscript{2}\thanks{Corresponding Authors}, Yunming Ye\textsuperscript{1} \\
  \textsuperscript{1}Harbin Institute of Technology, Shenzhen, \textsuperscript{2}Peng Cheng Laboratory, \\
    \texttt{lianggt@pcl.ac.cn}, \texttt{23B951062@stu.hit.edu.cn} \\
    \texttt{\{zhangbaoquan, yeyunming\}@hit.edu.cn} \\
    \texttt{wenzhiyuan2012@gmail.com}
}

\begin{document}

\maketitle

\begin{figure}[!h]
  \centering
  \vspace{-5mm}
\includegraphics[width=\linewidth]{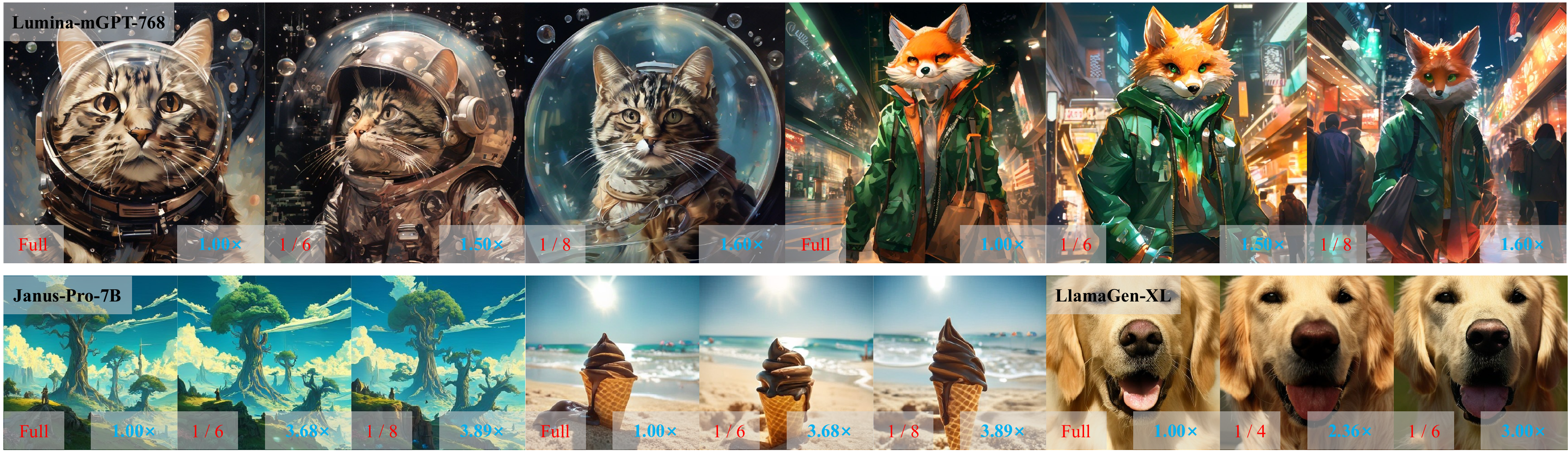}  
  \vspace{-5mm}
  \caption{
  HeadKV enables efficient autoregressive image generation by allocating asymmetric budgets across attention heads. With only 1/4, 1/6, and 1/8 of the KV cache retained, it achieves comparable visual quality to the full setting while significantly accelerating generation.
  }
  \label{fig:visualize}
  \vspace{-1mm}
\end{figure}

\begin{abstract}

Autoregressive (AR) visual generation has achieved remarkable performance but suffers from high memory usage and low throughput, as it requires caching previously generated visual tokens. Recent research has shown that retaining only a few lines of cache tokens can maintain high-quality images while significantly reducing memory usage and improving throughput. However, these methods allocate a fixed budget to each attention head, overlooking the heterogeneity among attention heads, leading to suboptimal memory allocation. 
In this paper, we observe that attention heads across different layers exhibit diverse attention patterns, where some heads focus on local neighborhoods while others capture broader contextual dependencies. Based on this insight, we propose a novel head-aware key-value (KV) cache compression framework for autoregressive image generation, called HeadKV, which assigns smaller budgets to locality-biased heads and larger budgets to heads with broader attention. A key challenge lies in identifying the type of each attention head to guide cache compression. We further observe that, within the same layer, each head exhibits consistent attention patterns across token positions, \emph{i.e.}, a head’s behavior for early tokens remains consistent with that for later tokens. This insight suggests that head types can be identified during the early stage and reused for KV compression throughout generation. Its advantage is that it requires no additional training or dataset-level statistics and generalizes seamlessly across different inputs. Moreover, we design a Stratified Token Eviction strategy to effectively preserve long-range information. Extensive experiments demonstrate its effectiveness across multiple autoregressive image generation models.
\end{abstract}

\section{Introduction}
Large Language Models (LLMs) based on the next-token prediction paradigm have achieved unparalleled success in the field of natural language processing, including text generation~\cite{achiam2023gpt,grattafiori2024llama}, code generation~\cite{wang2023review}, and translation~\cite{kleidermacher2026science}. Inspired by this success, researchers have extended such paradigms to the visual domain, leading to autoregressive visual generation models that represent images as sequences of visual tokens~\cite{chen2025janus,liu2024lumina-mgpt,sun2024autoregressive}. These models can generate high-resolution and photorealistic images by scaling the number of visual tokens, achieving resolutions of up to $1024 \times 1024$~\cite{liu2024lumina-mgpt}. However, this scaling introduces substantial memory and computational overhead due to key-value (KV) caching in autoregressive generation. For example, generating a $1024 \times 1024$ image requires over 50GB of GPU memory, posing significant challenges for practical deployment.

To mitigate the KV cache overhead, several recent works~\cite{cai2024pyramidkv,devoto2024simple,ge2023model} in natural language processing have explored KV cache management, such as token selection~\cite{li2024snapkv,zhang2023h2o}, cache eviction~\cite{feng2024ada,feng2025identify,fu2024not}, and cache quantization~\cite{hooper2024kvquant,yao2022zeroquant}. However, directly applying these techniques to the visual domain is non-trivial due to the inherent spatial structure of images. Recent work has begun to explore KV cache reduction specifically for autoregressive image generation. For example, LineAR~\cite{qin2025lineAR} shows that each generated token mainly focuses on its spatially adjacent tokens and initial conditional tokens, suggesting that only a small subset of tokens needs to be cached.

\begin{figure}
    \centering
    \includegraphics[width=1\linewidth]{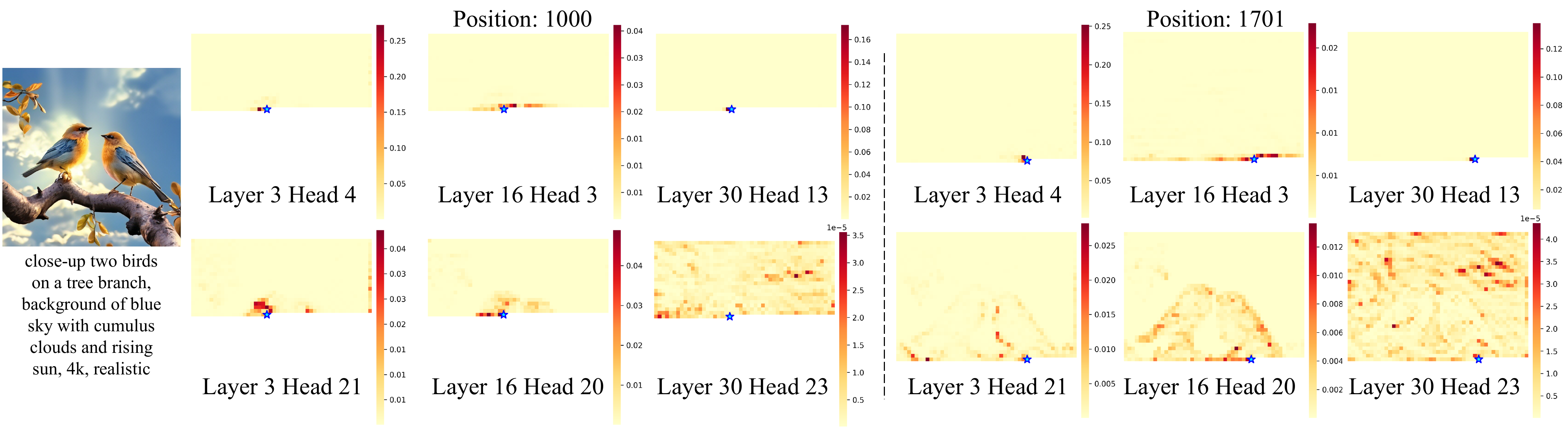}
    \vspace{-3mm}
    \caption{Visualizing the visual token attention map of the Lumina-mGPT-768 model. The left shows the input text and generated image, while the right shows attention maps from selected heads at layer-3, -16, and -30 at different token positions, revealing two patterns: local and global attention.}
    \label{fig:motivation}
    \vspace{-3mm}
\end{figure}
However, LineAR~\cite{qin2025lineAR} adopts a uniform token allocation strategy for attention heads, ignoring their inherent heterogeneity, which leads to suboptimal memory utilization. As illustrated in Fig.~\ref{fig:motivation}, we analyze the attention maps of different heads in the Lumina-mGPT-786 model~\cite{liu2024lumina-mgpt} and identify two distinct attention patterns: 
1) \textbf{Local heads.} Some heads (\emph{e.g.}, H3, H4, and H13) primarily attend to local regions. For these heads, caching a small set of nearby tokens is sufficient, and simple sliding-window mechanisms can effectively maintain the required context, without resorting to explicit token selection~\cite{li2024snapkv} or token eviction strategies~\cite{feng2024ada}; 
2) \textbf{Global heads}. In contrast, other heads (\emph{e.g.}, H20, H21, and H23) capture long-range dependencies, such as object structures and background regions. These heads require a substantially larger memory budget, where selectively preserving informative long-range tokens becomes critical, rather than merely retaining spatially adjacent tokens. This observation suggests that memory allocation should be adaptive to the attention patterns of individual heads, rather than being uniformly assigned.

Based on the above analysis, a natural strategy is to allocate a smaller budget to neighborhood-focused heads and a larger budget to those heads capturing long-range dependencies.
However, this raises a practical challenge: such an adaptive allocation requires identifying head types beforehand to guide KV cache compression.
This is non-trivial, as head behaviors are input-dependent. For instance, a specific head (\emph{e.g.}, Layer 3, Head 4) may exhibit local attention patterns for one input, while capturing more global dependencies for another. As a result, head types cannot be reliably determined beforehand through offline statistics or pre-defined rules.
Fortunately, we make another key observation: within a given input, each head exhibits consistent attention patterns across different token positions within the same layer. As illustrated in Fig.~\ref{fig:motivation}, the same head at different positions (\emph{e.g.}, positions 1000 and 1071) shows highly similar attention behaviors. This implies that, although head types vary across inputs, they remain largely invariant across token positions for a fixed input. Based on this property, we can identify head types during an early stage of generation, and then reuse this classification to guide KV cache compression throughout the subsequent decoding process. Notably, this approach requires neither additional training nor dataset-level statistics, and generalizes across diverse inputs.

As a result, we propose a novel training-free \emph{Head-aware KV} cache compression pipeline, termed HeadKV, to accelerate autoregressive image generation. Specifically, before compressing, HeadKV first determines head types using a threshold-based attention accumulation criterion. Following this, HeadKV assigns a compact budget to local heads and updates cache based on a sliding-windows strategy, while introducing a Stratified Token Eviction strategy for global heads to preserve more informative long-range historical tokens. The contributions of our paper include the following:
\begin{itemize}
    \item We conduct an in-depth analysis of attention heads in AR image generation models and identify two distinct attention patterns, \emph{i.e.}, local and global heads. Moreover, we reveal that head behaviors are invariant across token positions, providing a key insight for head-aware KV cache compression.
    \item Building on this insight, we propose a novel training-free head-aware KV cache compression framework, termed HeadKV, which allocates asymmetric budgets across attention heads by assigning compact budgets to local heads and larger budgets to global heads. For global heads, we further design a Stratified Token Eviction (STE) strategy to effectively preserve informative long-range dependencies.
    \item Extensive experiments demonstrate that HeadKV can significantly reduce KV cache memory usage and improve inference throughput in existing AR image generation models while maintaining high image quality.
\end{itemize}

\section{Related Works}
\subsection{Autoregressive visual generation}
Autoregressive (AR) visual generation typically represents images as sequences of discrete tokens obtained via vector quantization~\cite{esser2021taming}, and models their dependencies using Transformer-based architectures~\cite{ashish2017attention}. Building on this paradigm, prior works~\cite{razavi2019generating,ramesh2021zero,liang2024lg,liang2025towards} have significantly improved the quality and resolution of generated images. For instance, DALL$\cdot$E~\cite{ramesh2021zero} demonstrates that modeling images as token sequences and leveraging large-scale Transformer training enables strong text-to-image generation performance.
More recently, researchers have explored scaling up AR visual generation by adopting large language model (LLM) architectures~\cite{touvron2023llama,achiam2023gpt}, motivated by the success of scaling laws in language modeling. LlamaGen~\cite{grattafiori2024llama} directly adopts the LLaMA~\cite{touvron2023llama} architecture as its backbone and shows that increasing model size (from 111M to 3B parameters) leads to consistent improvements, indicating that scaling laws also apply to image generation. Lumina-mGPT~\cite{liu2024lumina-mgpt} further unifies multimodal generation by modeling both visual and textual tokens within a single autoregressive framework, while Janus-Pro~\cite{chen2025janus} introduces a decoupled visual encoding and generation strategy to improve both efficiency and generation quality. 
Despite these advances, AR visual generation models still suffer from substantial memory overhead and slow inference, primarily due to the large key-value (KV) cache required during autoregressive decoding. To address this, we focus on reducing the KV cache footprint and propose a novel HeadKV framework, which improves both memory efficiency and inference speed without sacrificing generation quality.

\subsection{Key-Value Cache Compression}
Key-Value (KV) cache compression aims to improve inference efficiency in LLMs by evicting less important key-value pairs while preserving generation quality. A core challenge lies in designing effective token selection strategies, \emph{i.e.}, selecting which tokens to retain or discard during inference. To address this, various eviction strategies~\cite{cai2024pyramidkv,feng2024ada,park2025keydiff} have been proposed. FastGen~\cite{ge2023model} identifies several common attention patterns and designs pattern-specific retention strategies. SnapKV~\cite{li2024snapkv} simplifies this process by selecting tokens based on attention-derived importance scores, showing that only a subset of tokens contributes significantly to generation quality. StreamingLLM~\cite{xiao2023efficient} identifies the attention sink phenomenon, where initial tokens play a critical role in maintaining performance, and retains these tokens together with recent context for efficient long-sequence modeling. H$_2$O~\cite{zhang2023h2o} leverages cumulative attention scores to identify and preserve high-impact tokens. Beyond eviction-based methods, other works~\cite{wan2024d2o,zhang2024cam,liu2024minicache} explore token merging strategies, which compress the KV cache by merging similar tokens instead of discarding them, to better preserve contextual information. Despite their success, these methods are tailored to LLMs and fail to consider the spatial semantics of images, leading to suboptimal performance. To address this, recent work, LineAR~\cite{qin2025lineAR}, proposes to manage the cache at the line level using a 2D view, preserving visual dependency regions while progressively evicting less informative tokens.

In this paper, we also focus on KV cache compression in the visual domain. Unlike LineAR~\cite{qin2025lineAR}, which allocates a fixed-size cache for all heads, we find that attention heads in visual AR models exhibit two distinct patterns: \emph{local heads} and \emph{global heads}. This heterogeneity suggests that a uniform budget for all heads is suboptimal. To this end, we propose a novel HeadKV framework, which allocates asymmetric budgets based on head attention patterns to accelerate AR visual generation, \emph{i.e.}, assigning smaller budgets to local heads and larger budgets to global heads, thereby improving both efficiency and generation quality. 
Our work is related to prior efforts such as DuoAttention~\cite{xiao2024duoattention} and Head-Level KV~\cite{fu2024not}, which also exploit head-level information for KV cache compression. However, these methods typically require additional training with synthetic or auxiliary data to identify specialized attention heads (\emph{e.g.}, retrieval heads), which may limit their generalizability and introduce extra computational overhead. In contrast, our approach directly leverages intrinsic attention patterns in visual autoregressive models, without requiring additional training or external supervision, thereby improving efficiency and general applicability.

\section{Preliminaries}
\textbf{Autoregressive Visual Generation.} 
Autoregressive (AR) visual generation models~\cite{chen2025janus,liu2024lumina-mgpt,grattafiori2024llama} formulate image synthesis as a sequential prediction problem, where an image is represented as a sequence of visual tokens $\{x_1, x_2, \dots, x_N\}$, with $N = h \times w$ denoting the sequence length. Given a conditional input $c$ (\emph{e.g.}, class labels or text descriptions), AR models generate tokens sequentially, where each token is conditioned on previously generated tokens $p_{\theta}(x) = \prod_{i=1}^{N} p_{\theta}(x_i \mid c, x_{<i}).$
Each AR layer employs causal self-attention, which maps the input $X$ to query, key, and value representations $Q$, $K$, and $V$ using projection matrices $W_q$, $W_k$, and $W_v$. The output is computed as:
\begin{equation}
    O = \mathrm{Softmax} \left( \frac{QK^T}{\sqrt{d}} \right) V.
\end{equation}
\textbf{KV Cache.} During autoregressive generation, computing attention for each new token requires projecting all previous tokens into key and value representations, leading to redundant computation. To address this, KV cache stores the previously computed keys and values as $K = [k_c, k_1, k_2, \dots, k_{i-1}]$ and $V = [v_c, v_1, v_2, \dots, v_{i-1}]$. For a new token $x_i$, only its query, key, and value need to be computed, and the cache is updated as $K \leftarrow [K, k_i]$, $V \leftarrow [V, v_i]$. While KV caching effectively reduces redundant computation and accelerates inference, its memory usage grows linearly with the sequence length. To mitigate this issue, LineAR~\cite{qin2025lineAR} exploits the observation that each token mainly attends to its local visual dependency region, and thus retains only a subset of tokens while evicting less informative ones. However, LineAR adopts a uniform token budget across attention heads, which overlooks head-aware heterogeneity, leading to suboptimal memory allocation.
\begin{figure*}
    \centering
    \includegraphics[width=1\linewidth]{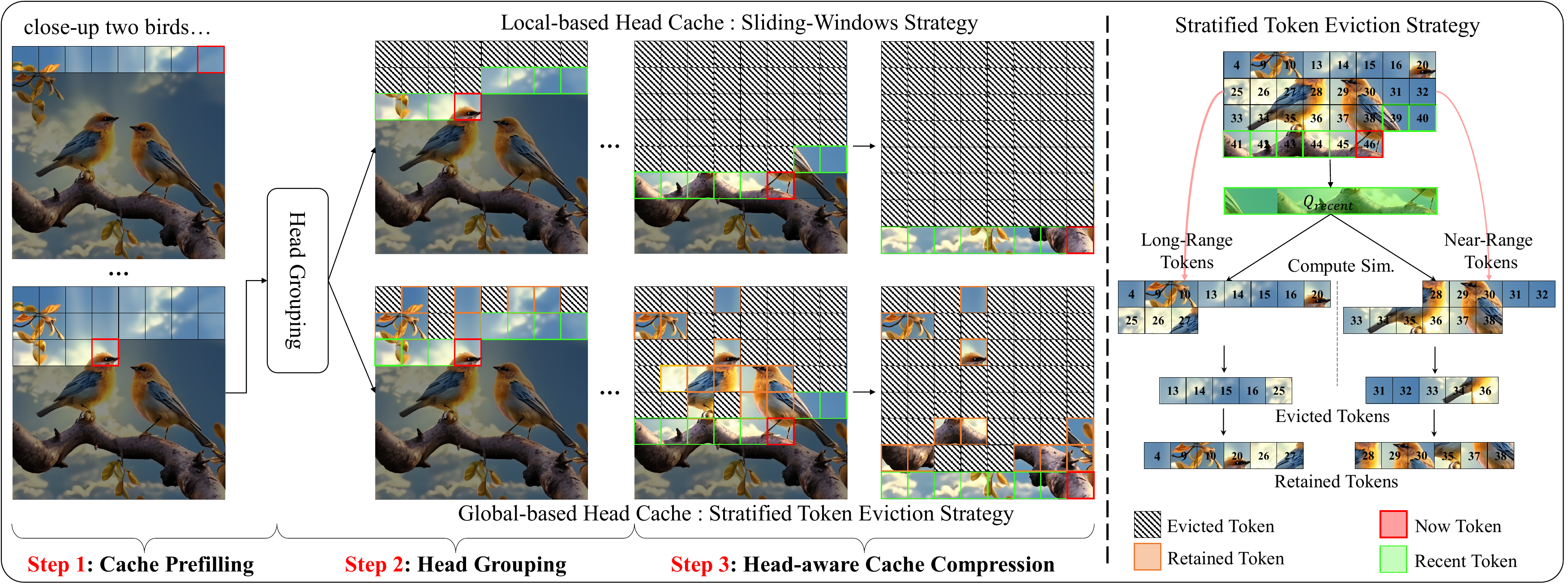}
    \caption{The illustration of our proposed HeadKV framework. The AR model first constructs an initial KV cache and generates visual tokens in an autoregressive manner. Once the cache size reaches a predefined threshold, a cumulative attention-based criterion is applied to classify attention heads into \emph{local} and \emph{global} types. Subsequently, the cache for local heads retains only the most recent tokens via a sliding-window mechanism, whereas the cache for global heads is updated using the proposed stratified token eviction strategy.}
    \label{fig:model}
\end{figure*}
\section{Proposed Method: HeadKV}
Existing works~\cite{qin2025lineAR} allocate a fixed KV cache budget to each attention head. However, as illustrated in Fig.~\ref{fig:motivation}, attention heads in AR models exhibit two distinct attention patterns: some focus on local neighborhoods, while others capture long-range dependencies, such as object structures or background regions. This heterogeneity indicates that memory allocation should be adaptive to the attention patterns of individual heads, rather than being uniformly assigned. To address this, we propose HeadKV, a novel framework that allocates asymmetric budgets based on head attention patterns for accelerating AR visual generative models. As illustrated in Fig~\ref{fig:model}, our HeadKV consists of three key steps, \emph{i.e.}, cache prefilling, head grouping, and head-aware cache compression. Next, we elaborate on each step in order:

\textbf{Step 1: \textit{Cache Prefilling}.} In this stage, the AR model encodes the conditional input $c$ and constructs an initial KV cache $K = [k_c]$ and $V = [v_c]$. It autoregressively generates visual tokens, updating the cache via $K \leftarrow [K, k_i]$ and $V \leftarrow [V, v_i]$. No compression is applied, as the accumulated tokens both form a stable image structure and provide sufficient context to estimate head-wise attention distributions for subsequent head type determination.

\textbf{Step 2: \textit{Head Grouping}.} When the cache size reaches a certain length, we need to identify the type of each head and leverage this classification to guide subsequent cache compression (\emph{i.e.}, assigning compact budgets to local heads and larger budgets to global-based heads). Fig~\ref{fig:motivation} illustrates that local heads tend to concentrate their attention on nearby tokens. Inspired by this observation, we propose a threshold-based attention accumulation criterion to identify head type. Specifically, for each head, we use the current token's query $Q$ to compute attention scores over the key cache $K$, and accumulate attention scores backward from the current position until a predefined threshold $\tau$ (set to 0.9) is reached, recording the minimum number of tokens required. If the required number of tokens is smaller than a predefined local window size, the head is classified as a \emph{local head}; otherwise, it is classified as a \emph{global head}.

\textbf{Step 3: \textit{Head-aware Cache Compression}.} After identifying head types, we apply different compression and update strategies based on their characteristics. To maintain a unified framework, we first formalize the structure of the KV cache and then describe how different heads manage their cache components within a fixed memory budget. Before compression, the KV caches of all heads are decomposed into three components:
\begin{equation}
K = [k_c, \mathcal{H}_k, k_{recent}], \quad
V = [v_c, \mathcal{H}_v, v_{recent}],
\end{equation}
where $k_c, v_c$ denote conditional tokens, $\mathcal{H}_k, \mathcal{H}_v$ represent historical tokens, \emph{i.e.}, tokens preceding the recent window but following the conditional tokens, and $k_{recent}, v_{recent}$ correspond to recent tokens. Note that the recent tokens form a fixed-size sliding window of length $w$. As new tokens are generated, they are appended to the recent set, while the oldest tokens are moved to the historical set.

\textit{Local Head.} As mentioned above, local heads primarily focus on the nearby context. Therefore, historical tokens can be omitted, and we set $\mathcal{H}_t = \emptyset$. The cache is fully allocated to recent tokens. To this end, the cache for local heads is as follows:
\begin{equation}
K_l = [k_c, \ k_{recent}], \quad
V_l = [v_c, \ k_{recent}].
\end{equation}
The cache updates are based on the standard sliding window mechanism.

\textit{Global Head.} In contrast, global heads exhibit a broader spatial context, where informative tokens may reside far from the current position. Retaining all historical tokens incurs significant memory and computational costs. To address this, we design a novel stratified token eviction strategy $f_{s}(\cdot)$ (see details below) that preserves more informative long-range tokens within the historical set, thereby preserving the quality of the generated image. To this end, the cache for global heads is as follows:
\begin{equation}
K_g \leftarrow [k_c, f_{s}(\mathcal{H}_k), \ k_{recent}], \quad
V_g \leftarrow [v_c, f_{s}(\mathcal{H}_v), \ v_{recent}].
\end{equation}

\textbf{Efficient Cache Update.} In practice, updating the cache at each decoding step incurs significant memory overhead due to frequent appending and eviction operations, leading to inefficient I/O usage. Following prior work~\cite{meng2025polaformer,qin2025lineAR,xiao2023efficient}, we adopt a periodic update strategy. Specifically, we buffer $P$ newly generated tokens and perform cache updates only when the buffer is full.

\textbf{Stratified Token Eviction.}
Previous works primarily select tokens based on importance scores~\cite{cai2024pyramidkv,li2024snapkv,zhang2023h2o}, while proposing various scoring strategies, such as multi-query aggregation~\cite{qin2025lineAR}, key feature-based metrics~\cite{park2025keydiff}, and L2-norm-based scoring~\cite{devoto2024simple}. However, in visual tasks, token correlations are typically dominated by strong local interactions, leading to a concentration of importance within nearby regions (Refer to the appendix's analysis of attention distribution). As a result, distant tokens, despite their semantic relevance, may be overshadowed by locally dominant tokens and are less likely to be selected.

To address this issue, as illustrated in Fig.~\ref{fig:model} (right), we propose a \emph{stratified token eviction} strategy. Specifically, we partition candidate tokens into two groups based on their distance from the current position, namely \emph{near-range} and \emph{long-range} tokens. Token selection is then performed independently within each group, preventing locally dominant tokens from dominating the selection process. This design encourages a more balanced cache allocation and improves the preservation of long-range dependencies.

Formally, given the historical token sets $\mathcal{H}_k, \mathcal{H}_v$ with length $T$, our goal is to evict $M$ tokens. We first partition the tokens into two disjoint subsets with a predefined ratio $r_s$ (default: $0.5$):
\begin{equation}
\mathcal{C}^{\text{long}} = [\mathcal{H}_k[:T r_s], \mathcal{H}_v[:T r_s]], \quad
\mathcal{C}^{\text{near}} = [\mathcal{H}_k[T r_s:], \mathcal{H}_v[T r_s:]].
\end{equation}
Following~\cite{qin2025lineAR}, we maintain a buffer of the past $P$ query features $Q_{\text{past}} = [q_{t-P}, \cdots, q_t]$, and compute importance scores for each subset:
\begin{equation}
\begin{aligned}
S^{\text{long}} &= \frac{1}{P} \mathbf{1}^\top 
\left(
Q_{\text{past}} (\mathcal{C}_k^{\text{long}})^\top / \sqrt{d}
\right), \\
S^{\text{near}} &= \frac{1}{P} \mathbf{1}^\top 
\left(
Q_{\text{past}} (\mathcal{C}_k^{\text{near}})^\top / \sqrt{d}
\right).
\end{aligned}
\end{equation}
We then select $M/2$ tokens with the lowest scores from each subset for eviction. Finally, the remaining tokens are concatenated to form the updated cache.\\
\textbf{Theoretical Analysis} We analyze Stratified Token Eviction (STE) from an information-theoretic resource allocation perspective. Our analysis shows that optimal KV cache allocation should prioritize tokens with high marginal information gain rather than relying solely on attention scores. Under the submodular information assumption, Global Top-K strategies tend to over-select locally redundant tokens, while STE improves information retention by encouraging diversity-aware allocation across different spatial domains. Please refer to the Appendix for details.

\begin{table*}[t]
\centering
\caption{Comparison of generation performance and computational efficiency on the GenEval~\cite{ghosh2023geneval} and DPG~\cite{hu2024dpg} benchmarks for Janus-Pro-1B, Janus-Pro-7B~\cite{chen2025janus}, Lumina-mGPT-768, and Lumina-mGPT-1024~\cite{liu2024lumina-mgpt}, evaluated under varying cache compression ratios $\rho$ with our method.}
\vspace{-3mm}
\label{tab:gen_dpg}

\footnotesize
\setlength{\tabcolsep}{3pt}
\renewcommand{\arraystretch}{1.05}

\newcommand{\graycell}[1]{%
  \begingroup
  \setlength{\fboxsep}{0pt}%
  \cellcolor{gray!30}#1%
  \endgroup
}

\resizebox{\textwidth}{!}{
\begin{tabular}{c c c cc ccccc cccc}
\bottomrule

\multirow{2}{*}{\textbf{Models}} 
& \multirow{2}{*}{$\boldsymbol{\rho}$} 
& \multirow{2}{*}{Method}
& \multicolumn{5}{c}{\textbf{GenEval} $\uparrow$} 
& \multicolumn{4}{c}{\textbf{DPG} $\uparrow$} \\

\cmidrule(lr){4-8} 
\cmidrule(lr){9-12}

&
& 
& \textbf{Single} 
& \textbf{Two} 
& \textbf{Count} 
& \textbf{Color} 
& \textbf{Overall} 
& \textbf{Entity} 
& \textbf{Relation} 
& \textbf{Attr.} 
& \textbf{Overall} \\

\midrule

\multirow{7}{*}{\thead{Janus-Pro-1B\\($N=576$)}}
& Full &
& 0.98 & 0.83 & 0.50 & 0.90 & 0.73 
& 86.58 & 92.26 & 87.83 & 82.21 \\

& \multirow{2}{*}{1/4} & LineAR
& 0.99            & 0.86          & 0.49           & 0.89           & 0.72 
& 88.30            & 92.88             & 87.06          & 82.16 \\

& & HeadKV
& \graycell{0.98} & \graycell{0.85} & \graycell{0.47} & \graycell{0.91} & \graycell{0.72} 
& \graycell{88.14} & \graycell{93.07} & \graycell{86.86} & \graycell{82.15} \\

& \multirow{2}{*}{1/6} & LineAR
& 0.99            & 0.81          & 0.51           & 0.89           & 0.71 
& 88.83           & 92.53             & 86.96          & 82.26 \\

& & HeadKV
& \graycell{0.98} & \graycell{0.82} & \graycell{0.44} & \graycell{0.90} & \graycell{0.71}
& \graycell{88.47} & \graycell{92.53} & \graycell{86.40} & \graycell{81.82} \\

& \multirow{2}{*}{1/8} & LineAR
& 0.98            & 0.80          & 0.42           & 0.87           & 0.68 
& 88.35            & 92.76             & 86.37          & 81.81 \\

& & HeadKV
& \graycell{0.98} & \graycell{0.81} & \graycell{0.43} & \graycell{0.87} & \graycell{0.70} 
& \graycell{87.73} & \graycell{90.47} & \graycell{85.08} & \graycell{81.93} \\

\midrule

\multirow{7}{*}{\thead{Janus-Pro-7B\\($N=576$)}} 
& Full &
& 0.99 & 0.86 & 0.57 & 0.91 & 0.79 
& 89.48 & 93.07 & 87.58 & 84.20 \\

& \multirow{2}{*}{1/4} & LineAR
& 0.99 & 0.89 & 0.58 & 0.93 & 0.80
&90.09    &93.26    &86.90 &84.06\\

& & HeadKV
& \graycell{0.98} & \graycell{0.88} & \graycell{0.57} & \graycell{0.89} & \graycell{0.78} 
& \graycell{89.03} & \graycell{93.23} & \graycell{87.24} & \graycell{83.92} \\

& \multirow{2}{*}{1/6} & LineAR
& 0.99 & 0.88 & 0.52 & 0.91 & 0.78 
&89.93       & 93.19     &87.26  & 83.76
 \\

& & HeadKV
& \graycell{0.98} & \graycell{0.88} & \graycell{0.57} & \graycell{0.90} & \graycell{0.77} 
& \graycell{89.35} & \graycell{93.23} & \graycell{87.24} & \graycell{83.73} \\

& \multirow{2}{*}{1/8} & LineAR
& 0.98 & 0.86 & 0.46 & 0.88 & 0.74 
&89.56       & 92.88     &87.10  & 83.82
 \\

& & HeadKV
& \graycell{0.99} & \graycell{0.87} & \graycell{0.52} & \graycell{0.89} & \graycell{0.76} 
& \graycell{89.87} & \graycell{93.03} & \graycell{87.10} & \graycell{83.78} \\
\midrule

\multirow{5}{*}{\thead{Lumina-mGPT-768 \\($N=2352$)}}
& Full &
& 1.00 & 0.73 & 0.25 & 0.82 & 0.52 
& 78.99 & 85.72 &  77.02 & 71.26 \\

& \multirow{2}{*}{1/6} & LineAR
& 0.99 & 0.74 & 0.28 & 0.84 & 0.52 
& 78.99  & 85.72 & 77.02 & 71.26 \\

&  & HeadKV 
& \graycell{0.99} & \graycell{0.74} & \graycell{0.29} & \graycell{0.85} & \graycell{0.52} 
& \graycell{79.64} & \graycell{86.07} & \graycell{76.97} & \graycell{71.30} \\

& \multirow{2}{*}{1/8} & LineAR
& 0.99 & 0.73 & 0.24 & 0.85 & 0.52 
& 80.38 & 86.59  & 78.05 & 71.55 \\

&  & HeadKV
& \graycell{0.98} & \graycell{0.75} & \graycell{0.24} & \graycell{0.82} & \graycell{0.51} 
& \graycell{80.28} & \graycell{86.67} & \graycell{77.51} & \graycell{71.33} \\

\midrule

\multirow{3}{*}{\thead{Lumina-mGPT-1024 \\($N=4160$)}}
& Full &
& 0.98 & 0.76 & 0.29 & 0.83 & 0.56 
& 86.67 & 90.95 & 84.42 & 79.98 \\

& \multirow{2}{*}{1/8} & LineAR
& 0.98  & 0.74 & 0.29 & 0.81  & 0.55  
& 86.67  & 90.91 & 84.22  & 79.65  \\

& & HeadKV
& \graycell{0.97}  & \graycell{0.73}  & \graycell{0.28} & \graycell{0.84} & \graycell{0.55} 
& \graycell{86.78} & \graycell{92.20} & \graycell{84.58}  & \graycell{80.05}  \\
\toprule
\end{tabular}
}
\vspace{-5mm}
\end{table*}

\section{Experiments}

\subsection{Experimental Setup}
\textbf{Backbone Models.}
We evaluate the effectiveness of our proposed method, \textit{HeadKV}, on a diverse set of autoregressive image generation backbones. Specifically, we consider four text-to-image models, including Janus-Pro-1B and Janus-Pro-7B~\cite{chen2025janus}, as well as Lumina-mGPT-768 and Lumina-mGPT-1024~\cite{liu2024lumina-mgpt}. In addition, we include LlamaGen-XL~\cite{grattafiori2024llama}, a class-conditional image generation model, to further demonstrate the generality of our approach across different generation paradigms. \\
\textbf{Evaluation Metrics.} For text-to-image models, we follow standard evaluation procedures on GenEval~\cite{ghosh2023geneval} and DPG~\cite{hu2024dpg} benchmarks, which provide diverse and challenging prompts to assess compositionality and semantic alignment. The evaluation protocols strictly follow the official settings of each benchmark.
In addition, we report Fréchet Inception Distance (FID)~\cite{heusel2017fid} scores for quantitative comparison. Specifically, we evaluate Janus-Pro-1B on the COCO-30K~\cite{lin2014microsoft} validation set using FID-30K. For the class-conditional setting, we evaluate LlamaGen-XL using FID-50K. \\
\textbf{Implementation Details.}
We introduce a key hyperparameter $\rho$ to control the cache compression ratio for global heads. Given a total of $N$ cached tokens, we retain $\rho N$ tokens for global heads. For example, when $\rho = 1/4$, only $N/4$ tokens are preserved. In contrast, local attention heads maintain only a small set of neighboring tokens, and their cache size is always smaller than that of global heads.
To handle the mismatch in sequence lengths between local and global heads, we explore three implementation strategies. First, we leverage variable-length attention support provided by the \texttt{flash\_attn\_varlen} library for efficient computation. Secondly, we pad the local attention tokens to match the global sequence length and apply a masking matrix to restrict the visible tokens for each local head. Thirdly, we calculate the attention independently within each group and then combine the results. In addition, we maintain a set of recent tokens, whose length is typically defined as two rows of image tokens, to preserve short-range dependencies during generation. All implementation details and hyperparameters follow the original configurations of the backbone models. More details can be found in the appendix.

\begin{minipage}[t]{.35\textwidth}
\centering
\footnotesize
\setlength{\tabcolsep}{4pt}
\renewcommand{\arraystretch}{1.00}
\newcommand{\graycell}[1]{%
  \begingroup
  \setlength{\fboxsep}{0pt}%
  \cellcolor{gray!30}#1%
  \endgroup
}
\captionof{table}{Results of class-to-image generation on \textbf{LlamaGen-XL}.}
\label{tab:llamagen}

\begin{tabular}{lccc}
\bottomrule
$\boldsymbol{\rho}$ & \textbf{Method}
& \textbf{FID-50K}$\downarrow$  & \textbf{IS}$\uparrow$ \\ \midrule

\textbf{Full} 
& - & 2.77 & 282.41 \\ \midrule

\multirow{2}{*}{$\rho = 1/4$} & LineAR 
& 2.64 & 279.95 \\
& HeadKV & 2.69 & 276.70 \\ \midrule

\multirow{2}{*}{$\rho = 1/6$}
& LineAR & 2.68& 277.04 \\
& HeadKV & 2.71 & 276.43 \\ \midrule

\multirow{2}{*}{$\rho = 1/8$}
& LineAR & 2.95 & 282.91 \\
& HeadKV & \graycell{2.75} & \graycell{273.14} \\

\toprule
\end{tabular}
\end{minipage}\qquad
\begin{minipage}[t]{.6\textwidth}
\centering
\footnotesize
\setlength{\tabcolsep}{1pt}
\renewcommand{\arraystretch}{1.00}

\captionof{table}{Quantitative comparison of different KV cache compression methods.}
\label{baselinecompar}

\begin{tabular}{lccccc}
\bottomrule
\multirow{2}{*}{\textbf{Method}} 
& \multicolumn{2}{c}{\textbf{LlamaGen-XL(FID-50K$\downarrow$)}} 
& \multicolumn{2}{c}{\textbf{Janus-Pro-1B(FID-30K$\downarrow$)}} \\
\cmidrule(lr){2-3} \cmidrule(lr){4-5}
& $\rho{=}1/6$ & $\rho{=}1/8$ 
& $\rho{=}1/6$ & $\rho{=}1/8$ \\
\midrule

\textbf{Full} & \multicolumn{2}{c}{2.77} & \multicolumn{2}{c}{22.05} \\

\midrule

Streaming~\cite{xiao2023efficient} & 3.09  & 4.37   & 24.80 & 32.17 \\
H2O~\cite{zhang2023h2o}       & 3.06  & 3.72   & 25.87 & 29.46 \\
R-KV~\cite{cai2025r}      & 2.77  & 3.45   & 24.63 & 25.55 \\
LineAR~\cite{qin2025lineAR}    & \textbf{2.68} & 2.95   & 22.46 & 24.03 \\

\midrule
\rowcolor{gray!30} HeadKV  & 2.71 & \textbf{2.75} & \textbf{21.49} & \textbf{22.41} \\

\toprule
\end{tabular}
\end{minipage}

\vspace{-3mm}
\subsection{Discussion of Results}
\begin{figure*}[t]
    \centering
    \includegraphics[width=1\linewidth]{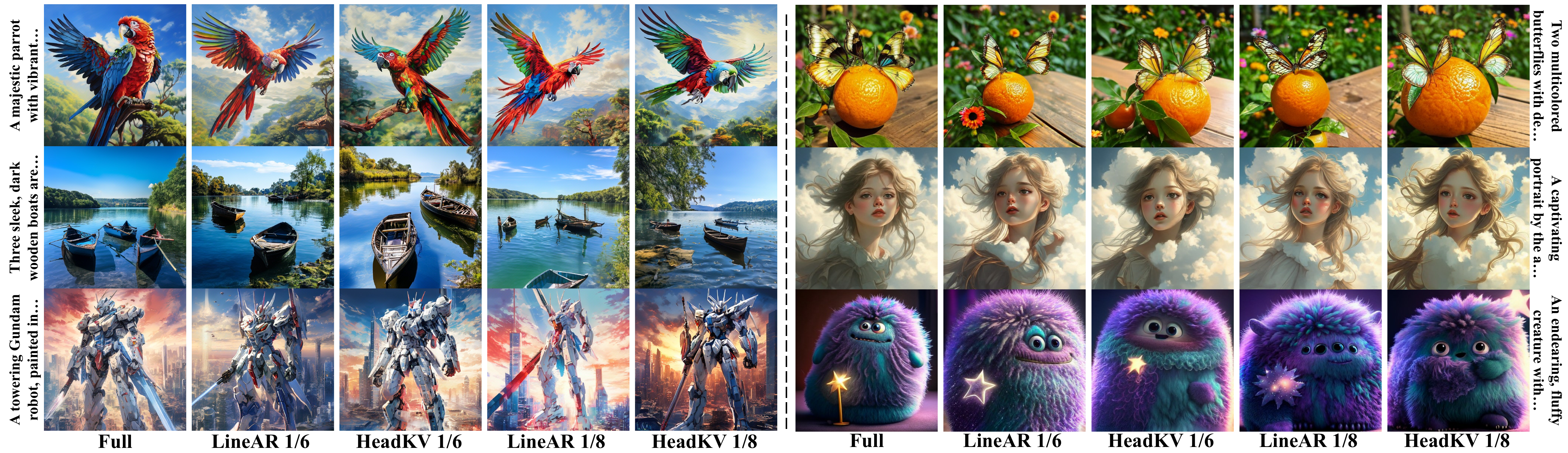}
    \vspace{-5mm}
    \caption{Qualitative comparison of text-to-image generation under different compression ratio $\rho$. The left is Lumina-mGPT-786, and right is Janus-Pro-7B.}
    \label{fig:t2i_comp}
\end{figure*}

\textbf{Text-to-Image Generation.} In Tab.~\ref{tab:gen_dpg}, we evaluate HeadKV on four text-to-image models using GenEval~\cite{ghosh2023geneval} and DPG~\cite{hu2024dpg} benchmarks, comparing against Full Cache and LineAR~\cite{qin2025lineAR} under different compression ratios. The results show several key conclusions. Firstly, HeadKV achieves performance comparable to the full-cache baseline while significantly improving efficiency (see Section~\ref{effic}), demonstrating the effectiveness of our approach. Secondly, compared to LineAR, HeadKV yields slightly lower performance across some metrics, mainly due to a more aggressive cache reduction for local heads. This trade-off yields reduced memory usage and improved inference speed. Third, HeadKV exhibits less performance degradation as the compression ratio increases. This is attributed to our stratified token eviction strategy, which better preserves long-range informative tokens under limited budgets, leading to more stable generation quality. Finally, we also provide a qualitative comparison in Fig.~\ref{fig:t2i_comp}, which further confirms that HeadKV preserves fine-grained details and produces visually coherent images. \\
\textbf{Class-to-Image Generation.} In Tab.~\ref{tab:llamagen}, we extend our method to the class-to-image generation model LlamaGen-XL. While LineAR achieves slightly better results under moderate compression (\emph{e.g.}, $\rho = 1/4$ and $1/6$), the performance gap remains relatively small. This suggests that allocating a uniform budget across all heads is actually redundant, validating the motivation behind our method.
Moreover, as the compression becomes more extreme ($\rho = 1/8$), HeadKV demonstrates a clear advantage. In particular, HeadKV achieves better FID (2.75 vs 2.95), while LineAR exhibits noticeable performance degradation. This indicates that HeadKV exhibits strong robustness under tight cache budgets, primarily due to our proposed stratified token eviction strategy. This strategy better preserves informative long-range dependencies. \\
\textbf{Comparison with Baselines.} Following the LineAR~\cite{qin2025lineAR}, we further compare our method with some representative KV cache compression methods proposed to accelerate large language models. We extend them to AR image generation, including models such as Streaming~\cite{xiao2023efficient}, which maintain a fixed cache size based on a sliding-windows strategy, H2O~\cite{zhang2023h2o}, which selects more important tokens based on accumulated attention scores, and R-KV~\cite{cai2025r}, which selects tokens based on contextual redundancy among reasoning contexts. Specifically, we evaluate LlamaGen-XL using FID-50K~\cite{heusel2017fid,liang2026improved} on Imagenet. For the Janus-pro-1B, we randomly select 30K samples from the COCO val dataset~\cite{lin2014microsoft} for evaluation. From the results, we observe that headKV achieves the best performance at a more aggressive compression ratio $\rho=1/8$, demonstrating the effectiveness of our proposed approach.

\vspace{-10pt}
\subsection{Ablation Study}
\vspace{-10pt}
\begin{table*}[htb]
\centering
\caption{Efficiency analysis on LlamaGen-XL, Janus-7B-Pro and Lumina-mGPT-768 models under different compressive ratio $\rho$, evaluated on NVIDIA A6000 GPUs.}
\vspace{-3mm}
\label{tab:efficient}
\footnotesize
\setlength{\tabcolsep}{3pt}
\renewcommand{\arraystretch}{1.00}
\resizebox{\textwidth}{!}{
\begin{tabular}{cccccccccc}
\bottomrule
\textbf{Models}                                   & \textbf{$\rho$}      & \textbf{Method} & \textbf{$B$} & \textbf{Batch} & \textbf{Memory} & \textbf{Mem. Saving} & \textbf{Latency} & \textbf{Through.} & \textbf{Speedup} \\ \midrule 
\multirow{5}{*}{\thead{LlamaGen-XL \\($N=576$)}}      & 1                    & Full            & 576          & 128            & 33.91GB         & -                    & 101.97s          & 1.26$\text{it/s}$ & 1.00$\times$     \\
                                                  & \multirow{2}{*}{1/4} & LineAR          & 144           & 128            & 14.88GB         & 56.11\%              & 44.95s           & 2.85$\text{it/s}$ & 2.26$\times$     \\
                                                  &                      & HeadKV          & 144           & 128            & 14.63GB         & 56.85\%              & 42.99s           & 2.98$\text{it/s}$ & 2.36$\times$     \\

                                                  & \multirow{2}{*}{1/6} & LineAR          & 96           & 128            & 12.62GB         & 62.78\%              & 34.61s           & 3.70$\text{it/s}$ & 2.93$\times$     \\
                                                  &                      & HeadKV          & 96           & 128            & 12.34GB         & 63.60\%              & 34.03s           & 3.78$\text{it/s}$ & 3.00$\times$     \\ \midrule

\multirow{5}{*}{\thead{Janus-Pro-7B\\($N=576$)}}      & 1                    & Full            & 576          & 55             & 45.04GB         & -                    & 144.44s          & 0.38$\text{it/s}$ & 1.00$\times$     \\
                                                  & \multirow{2}{*}{1/6} & LineAR          & 96           & 55             & 22.03GB         & 51.08\%              & 39.27s           & 1.40$\text{it/s}$ & 3.68$\times$     \\
                                                  &                      & HeadKV          & 96           & 55             & 21.58GB         & 52.08\%              & 39.14s           & 1.40$\text{it/s}$ & 3.68$\times$     \\
                                                  & \multirow{2}{*}{1/8} & LineAR          & 72           & 55             & 20.66GB         & 54.12\%              & 35.30s           & 1.55$\text{it/s}$ & 4.07$\times$     \\
                                                  &                      & HeadKV          & 72           & 55             & 20.45GB         & 54.59\%              & 36.95s           & 1.48$\text{it/s}$ & 3.89$\times$     \\ \midrule 
\multirow{3}{*}{\thead{Lumina-mGPT-768 \\($N=2352$)}} & 1                    & Full            & 2352         & 6             & 35.27GB         & -                    & 287.46s          & 0.020$\text{it/s}$ & 1.00$\times$     \\
                                                  & \multirow{2}{*}{1/8} & LineAR          &  294          &  6              &  17.31GB               & 50.92\%                     &  179.38s               &  0.033$\text{it/s}$                 & 1.65$\times$                 \\
                                                  &                      & HeadKV          &  294           &  6             &  17.21GB               & 51.20\%                     & 218.94s                 &  0.027$\text{it/s}$                 & 1.35$\times$                 \\ \toprule 
\end{tabular}
}
\vspace{-5mm}
\end{table*}
\textbf{Efficiency Analysis} \label{effic} In Tab.~\ref{tab:efficient}, we present a comprehensive efficiency analysis in terms of memory usage and inference speed. Due to the heterogeneous sequence lengths introduced by HeadKV, we consider three implementation strategies for attention computation: (1) a variable-length implementation based on \texttt{flash\_attn\_varlen}, (2) a padding-based strategy with masking, and (3) an independent attention computation. We report the best results of implementation in Tab.~\ref{tab:efficient}. From the results, we observe that: 
1) HeadKV achieves consistently low memory usage across all models and compression ratios, primarily due to our more aggressive compression strategy for local heads. In particular, this strategy significantly improves inference speed for LlamaGen-XL and Janus-Pro-7B, surpassing LineAR, demonstrating the effectiveness of our method.
2) On larger models such as Lumina-mGPT-768, HeadKV exhibits slightly higher latency compared to LineAR. This is mainly due to the inefficient cache updates caused by grouping, leading to latency. To further evaluate practical deployment efficiency, we provide results of the padding-based strategy in the Appendix. This implementation achieves comparable inference speed to LineAR. \\
\textbf{Partition ratio $r_{s}$ of stratified token eviction} 
As mentioned above, we propose a novel stratified token eviction strategy that introduces a partitioning ratio hyperparameter, $r_s$, to control the number of long-range tokens. In Fig.~\ref{fig:r_s}, we analyze the impact of this hyperparameter on performance. 
The results show a clear non-monotonic relationship between $r_s$ and performance. When $r_s$ is too small (\emph{e.g.}, $r_s=0.2$), the model allocates insufficient budget to long-range tokens, leading to degraded performance due to the loss of global contextual information. As $r_s$ increases, the performance improves and reaches its optimal region around $r_s \in [0.5, 0.6]$. 
However, when $r_s$ becomes too large (e.g., $r_s=0.8$), the performance deteriorates again, indicating that over-allocating budget to long-range tokens weakens the model’s ability to capture fine-grained local dependencies. 
Overall, these results suggest that a balanced allocation between near-range and long-range tokens is crucial for maintaining both local detail fidelity and global structural coherence. \\
\textbf{Is our threshold-based grouping strategy effective?} 
Based on observed attention patterns, we propose a threshold-based attention accumulation criterion to identify attention head types. To evaluate the effectiveness of this strategy, we conduct an accuracy analysis. Specifically, we evaluate the Janus-Pro-1B model on 10 input samples from the DPG dataset. We first obtain head type classifications at an early stage as a reference, and then compare them with classifications obtained at later stages. The results are summarized in Tab.~\ref{tab:group_acc}. The results indicate that our threshold-based classification strategy achieves high accuracy, supporting the observation that once an attention head exhibits a local or global tendency in the early stage, this tendency remains stable throughout the inference process. In Tab.~\ref{tab:group_acc}, $L \rightarrow G$ denotes the misclassification rate where a global head is incorrectly identified as a local head. This error rate is relatively low, around 7\%–8\%. This error does not significantly affect overall model performance, likely due to the high redundancy in visual tokens. 
\vspace{-10pt}

\begin{minipage}[t]{.40\textwidth}
\centering
\captionof{figure}{Results of hyperparameter partitioning ratio $r_s$.}
\includegraphics[width=1\linewidth]{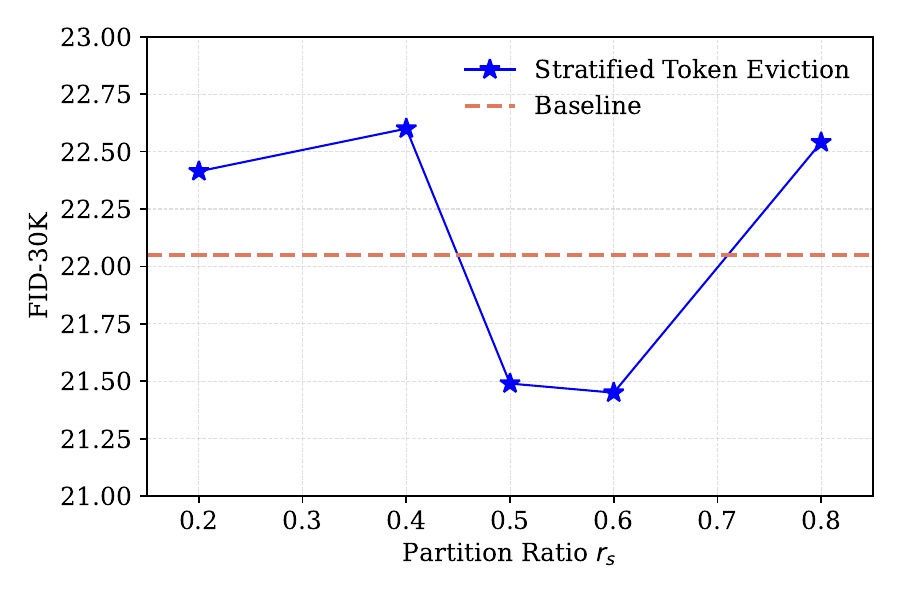}
\label{fig:r_s}
\end{minipage}\qquad
\begin{minipage}[t]{.55\textwidth}
\centering
\footnotesize
\setlength{\tabcolsep}{1pt}
\renewcommand{\arraystretch}{1.15}
\captionof{table}{Results of classification accuracy and misclassification accuracy at different thresholds. The $L \rightarrow G$ denotes the frequency of misclassification where a truly global head is incorrectly identified as a local head.}
\label{tab:group_acc}
\begin{tabular}{cccc}
\bottomrule
Threshold & Total Acc.$\uparrow$ & L $\rightarrow$ G $\downarrow$ & FID-30K$\downarrow$ \\ \hline
0.85      & 88.01      & 8.70\%            &  21.50       \\
0.90      & 88.78      & 8.45\%            &  21.49       \\
0.95      & 90.84      & 7.27\%            &  21.45       \\ \toprule
\end{tabular}
\end{minipage}

\vspace{-15pt}
\section{Conclusion}
In this paper, we propose a novel training-free, head-aware KV cache compression framework for accelerating autoregressive image generation by allocating different budgets based on attention head patterns. In particular, we identify two distinct types of attention heads, \emph{i.e.}, local and global heads, and observe that once an attention head exhibits a local or global pattern in the early stage, this behavior remains stable throughout the inference process. Moreover, we design a stratified token eviction strategy to effectively preserve long-range dependencies. Extensive experiments validate the effectiveness and versatility of our approach across diverse backbone models.
\newpage


\appendix
\bibliographystyle{plain}
\bibliography{ref}

\begin{thebibliography}{10}

\bibitem{achiam2023gpt}
Josh Achiam, Steven Adler, Sandhini Agarwal, Lama Ahmad, Ilge Akkaya, Florencia~Leoni Aleman, Diogo Almeida, Janko Altenschmidt, Sam Altman, Shyamal Anadkat, et~al.
\newblock Gpt-4 technical report.
\newblock {\em arXiv preprint arXiv:2303.08774}, 2023.

\bibitem{ashish2017attention}
Vaswani Ashish.
\newblock Attention is all you need.
\newblock {\em Advances in neural information processing systems}, 30:I, 2017.

\bibitem{cai2025r}
Zefan Cai, Wen Xiao, Hanshi Sun, Cheng Luo, Yikai Zhang, Ke~Wan, Yucheng Li, Yeyang Zhou, Li-Wen Chang, Jiuxiang Gu, et~al.
\newblock R-kv: Redundancy-aware kv cache compression for training-free reasoning models acceleration.
\newblock {\em arXiv e-prints}, pages arXiv--2505, 2025.

\bibitem{cai2024pyramidkv}
Zefan Cai, Yichi Zhang, Bofei Gao, Yuliang Liu, Yucheng Li, Tianyu Liu, Keming Lu, Wayne Xiong, Yue Dong, Junjie Hu, et~al.
\newblock Pyramidkv: Dynamic kv cache compression based on pyramidal information funneling.
\newblock {\em arXiv preprint arXiv:2406.02069}, 2024.

\bibitem{chen2025janus}
Xiaokang Chen, Zhiyu Wu, Xingchao Liu, Zizheng Pan, Wen Liu, Zhenda Xie, Xingkai Yu, and Chong Ruan.
\newblock Janus-pro: Unified multimodal understanding and generation with data and model scaling.
\newblock {\em arXiv preprint arXiv:2501.17811}, 2025.

\bibitem{devoto2024simple}
Alessio Devoto, Yu~Zhao, Simone Scardapane, and Pasquale Minervini.
\newblock A simple and effective l\_2 norm-based strategy for kv cache compression.
\newblock In {\em Proceedings of the 2024 Conference on Empirical Methods in Natural Language Processing}, pages 18476--18499, 2024.

\bibitem{esser2021taming}
Patrick Esser, Robin Rombach, and Bjorn Ommer.
\newblock Taming transformers for high-resolution image synthesis.
\newblock In {\em Proceedings of the IEEE/CVF conference on computer vision and pattern recognition}, pages 12873--12883, 2021.

\bibitem{feng2024ada}
Yuan Feng, Junlin Lv, Yukun Cao, Xike Xie, and S~Kevin Zhou.
\newblock Ada-kv: Optimizing kv cache eviction by adaptive budget allocation for efficient llm inference.
\newblock {\em arXiv preprint arXiv:2407.11550}, 2024.

\bibitem{feng2025identify}
Yuan Feng, Junlin Lv, Yukun Cao, Xike Xie, and S~Kevin Zhou.
\newblock Identify critical kv cache in llm inference from an output perturbation perspective.
\newblock {\em arXiv preprint arXiv:2502.03805}, 2025.

\bibitem{fu2024not}
Yu~Fu, Zefan Cai, Abedelkadir Asi, Wayne Xiong, Yue Dong, and Wen Xiao.
\newblock Not all heads matter: A head-level kv cache compression method with integrated retrieval and reasoning.
\newblock {\em arXiv preprint arXiv:2410.19258}, 2024.

\bibitem{ge2023model}
Suyu Ge, Yunan Zhang, Liyuan Liu, Minjia Zhang, Jiawei Han, and Jianfeng Gao.
\newblock Model tells you what to discard: Adaptive kv cache compression for llms.
\newblock {\em arXiv preprint arXiv:2310.01801}, 2023.

\bibitem{ghosh2023geneval}
Dhruba Ghosh, Hannaneh Hajishirzi, and Ludwig Schmidt.
\newblock Geneval: An object-focused framework for evaluating text-to-image alignment.
\newblock {\em Advances in Neural Information Processing Systems}, 36:52132--52152, 2023.

\bibitem{grattafiori2024llama}
Aaron Grattafiori, Abhimanyu Dubey, Abhinav Jauhri, Abhinav Pandey, Abhishek Kadian, Ahmad Al-Dahle, Aiesha Letman, Akhil Mathur, Alan Schelten, Alex Vaughan, et~al.
\newblock The llama 3 herd of models.
\newblock {\em arXiv preprint arXiv:2407.21783}, 2024.

\bibitem{heusel2017fid}
Martin Heusel, Hubert Ramsauer, Thomas Unterthiner, Bernhard Nessler, and Sepp Hochreiter.
\newblock Gans trained by a two time-scale update rule converge to a local nash equilibrium.
\newblock {\em Advances in neural information processing systems}, 30, 2017.

\bibitem{hooper2024kvquant}
Coleman Hooper, Sehoon Kim, Hiva Mohammadzadeh, Michael~W Mahoney, Yakun~S Shao, Kurt Keutzer, and Amir Gholami.
\newblock Kvquant: Towards 10 million context length llm inference with kv cache quantization.
\newblock {\em Advances in Neural Information Processing Systems}, 37:1270--1303, 2024.

\bibitem{hu2024dpg}
Xiwei Hu, Rui Wang, Yixiao Fang, Bin Fu, Pei Cheng, and Gang Yu.
\newblock Ella: Equip diffusion models with llm for enhanced semantic alignment.
\newblock {\em arXiv preprint arXiv:2403.05135}, 2024.

\bibitem{kleidermacher2026science}
Hannah~Calzi Kleidermacher and James Zou.
\newblock Science across languages: assessing llm multilingual translation of scientific papers.
\newblock In {\em Findings of the Association for Computational Linguistics: EACL 2026}, pages 3932--3947, 2026.

\bibitem{li2024snapkv}
Yuhong Li, Yingbing Huang, Bowen Yang, Bharat Venkitesh, Acyr Locatelli, Hanchen Ye, Tianle Cai, Patrick Lewis, and Deming Chen.
\newblock Snapkv: Llm knows what you are looking for before generation.
\newblock {\em Advances in Neural Information Processing Systems}, 37:22947--22970, 2024.

\bibitem{liang2024lg}
Guotao Liang, Baoquan Zhang, Yaowei Wang, Xutao Li, Yunming Ye, Huaibin Wang, Chuyao Luo, Kola Ye, and Linfeng Luo.
\newblock Lg-vq: Language-guided codebook learning.
\newblock {\em Advances in Neural Information Processing Systems}, 37:139700--139724, 2024.

\bibitem{liang2026improved}
Guotao Liang, Baoquan Zhang, Zhiyuan Wen, Zihao Han, and Yunming Ye.
\newblock Improved masked image generation with knowledge-augmented token representations.
\newblock In {\em Proceedings of the AAAI Conference on Artificial Intelligence}, pages 6817--6825, 2026.

\bibitem{liang2025towards}
Guotao Liang, Baoquan Zhang, Zhiyuan Wen, Junteng Zhao, Yunming Ye, Kola Ye, and Yao He.
\newblock Towards improved text-aligned codebook learning: Multi-hierarchical codebook-text alignment with long text.
\newblock In {\em Proceedings of the Computer Vision and Pattern Recognition Conference}, pages 4060--4069, 2025.

\bibitem{lin2014microsoft}
Tsung-Yi Lin, Michael Maire, Serge Belongie, James Hays, Pietro Perona, Deva Ramanan, Piotr Doll{\'a}r, and C~Lawrence Zitnick.
\newblock Microsoft coco: Common objects in context.
\newblock In {\em European conference on computer vision}, pages 740--755. Springer, 2014.

\bibitem{liu2024minicache}
Akide Liu, Jing Liu, Zizheng Pan, Yefei He, Gholamreza Haffari, and Bohan Zhuang.
\newblock Minicache: Kv cache compression in depth dimension for large language models.
\newblock {\em Advances in Neural Information Processing Systems}, 37:139997--140031, 2024.

\bibitem{liu2024lumina-mgpt}
Dongyang Liu, Shitian Zhao, Le~Zhuo, Weifeng Lin, Yu~Qiao, Hongsheng Li, and Peng Gao.
\newblock Lumina-mgpt: Illuminate flexible photorealistic text-to-image generation with multimodal generative pretraining, 2024.

\bibitem{meng2025polaformer}
Weikang Meng, Yadan Luo, Xin Li, Dongmei Jiang, and Zheng Zhang.
\newblock Polaformer: Polarity-aware linear attention for vision transformers.
\newblock {\em arXiv preprint arXiv:2501.15061}, 2025.

\bibitem{park2025keydiff}
Junyoung Park, Dalton Jones, Matthew~J Morse, Raghavv Goel, Mingu Lee, and Chris Lott.
\newblock Keydiff: Key similarity-based kv cache eviction for long-context llm inference in resource-constrained environments.
\newblock {\em arXiv preprint arXiv:2504.15364}, 2025.

\bibitem{qin2025lineAR}
Ziran Qin, Youru Lv, Mingbao Lin, Zeren Zhang, Chanfan Gan, Tieyuan Chen, and Weiyao Lin.
\newblock Autoregressive image generation needs only a few lines of cached tokens.
\newblock {\em arXiv preprint arXiv:2512.04857}, 2025.

\bibitem{ramesh2021zero}
Aditya Ramesh, Mikhail Pavlov, Gabriel Goh, Scott Gray, Chelsea Voss, Alec Radford, Mark Chen, and Ilya Sutskever.
\newblock Zero-shot text-to-image generation.
\newblock In {\em International conference on machine learning}, pages 8821--8831. Pmlr, 2021.

\bibitem{razavi2019generating}
Ali Razavi, Aaron Van~den Oord, and Oriol Vinyals.
\newblock Generating diverse high-fidelity images with vq-vae-2.
\newblock {\em Advances in neural information processing systems}, 32, 2019.

\bibitem{so2025grouped}
Junhyuk So, Juncheol Shin, Hyunho Kook, and Eunhyeok Park.
\newblock Grouped speculative decoding for autoregressive image generation.
\newblock In {\em Proceedings of the IEEE/CVF International Conference on Computer Vision}, pages 15375--15384, 2025.

\bibitem{sun2024autoregressive}
Peize Sun, Yi~Jiang, Shoufa Chen, Shilong Zhang, Bingyue Peng, Ping Luo, and Zehuan Yuan.
\newblock Autoregressive model beats diffusion: Llama for scalable image generation.
\newblock {\em arXiv preprint arXiv:2406.06525}, 2024.

\bibitem{touvron2023llama}
Hugo Touvron, Thibaut Lavril, Gautier Izacard, Xavier Martinet, Marie-Anne Lachaux, Timoth{\'e}e Lacroix, Baptiste Rozi{\`e}re, Naman Goyal, Eric Hambro, Faisal Azhar, et~al.
\newblock Llama: Open and efficient foundation language models.
\newblock {\em arXiv preprint arXiv:2302.13971}, 2023.

\bibitem{wan2024d2o}
Zhongwei Wan, Xinjian Wu, Yu~Zhang, Yi~Xin, Chaofan Tao, Zhihong Zhu, Xin Wang, Siqi Luo, Jing Xiong, Longyue Wang, et~al.
\newblock D2o: Dynamic discriminative operations for efficient long-context inference of large language models.
\newblock {\em arXiv preprint arXiv:2406.13035}, 2024.

\bibitem{wang2023review}
Jianxun Wang and Yixiang Chen.
\newblock A review on code generation with llms: Application and evaluation.
\newblock In {\em 2023 IEEE International Conference on Medical Artificial Intelligence (MedAI)}, pages 284--289. IEEE, 2023.

\bibitem{xiao2024duoattention}
Guangxuan Xiao, Jiaming Tang, Jingwei Zuo, Junxian Guo, Shang Yang, Haotian Tang, Yao Fu, and Song Han.
\newblock Duoattention: Efficient long-context llm inference with retrieval and streaming heads.
\newblock {\em arXiv preprint arXiv:2410.10819}, 2024.

\bibitem{xiao2023efficient}
Guangxuan Xiao, Yuandong Tian, Beidi Chen, Song Han, and Mike Lewis.
\newblock Efficient streaming language models with attention sinks.
\newblock {\em arXiv preprint arXiv:2309.17453}, 2023.

\bibitem{yao2022zeroquant}
Zhewei Yao, Reza Yazdani~Aminabadi, Minjia Zhang, Xiaoxia Wu, Conglong Li, and Yuxiong He.
\newblock Zeroquant: Efficient and affordable post-training quantization for large-scale transformers.
\newblock {\em Advances in neural information processing systems}, 35:27168--27183, 2022.

\bibitem{zhang2024cam}
Yuxin Zhang, Yuxuan Du, Gen Luo, Yunshan Zhong, Zhenyu Zhang, Shiwei Liu, and Rongrong Ji.
\newblock Cam: Cache merging for memory-efficient llms inference.
\newblock In {\em Forty-first international conference on machine learning}, 2024.

\bibitem{zhang2023h2o}
Zhenyu Zhang, Ying Sheng, Tianyi Zhou, Tianlong Chen, Lianmin Zheng, Ruisi Cai, Zhao Song, Yuandong Tian, Christopher R{\'e}, Clark Barrett, et~al.
\newblock H2o: Heavy-hitter oracle for efficient generative inference of large language models.
\newblock {\em Advances in Neural Information Processing Systems}, 36:34661--34710, 2023.

\end{thebibliography}

\newpage
\section{Appendix}

\subsection{Limitation and Future Work.}
While HeadKV achieves strong acceleration with minimal quality degradation, its efficiency is not yet optimal due to additional overhead in attention mask management and head-wise budget allocation. Future work will explore whether similar attention patterns exist in large language models (LLMs), and investigate extending the head-aware budget allocation strategy to LLMs for further efficiency gains.

\subsection{More Implementation Details.}
Regarding cache management, we apply head grouping once the cache length exceeds a predefined threshold. This threshold is set to 100 for Janus-Pro and LlamaGen, 300 for Lumina-mGPT-768, and 500 for Lumina-mGPT-1024. Hyperparameters used for inference, such as CFG and Topk sampling, follow the default hyperparameters of the backbone model. Specifically, for DPG on Lumina-mGPT-768, we set CFG=2 for all models. 
    
\subsection{Efficiency Analysis}
In Tab.~\ref{tab:efficient}, we provide an experiment of efficiency analysis. This result shows that this strategy can significantly improve the speedup on the LlamaGen-XL model, but the speedup is not good on larger models, mainly due to inefficient cache updates caused by grouping. In this section, we present a padding-based implementation strategy that pads the local head to match the global head's length and introduces a masked attention constraint on the visible length of the local head. The results are shown in Tab~\ref{tab:efficient_padding}. The results show that our method achieves comparable acceleration performance to the linear baseline. The slight overhead mainly comes from maintaining an additional attention mask matrix, which offsets part of the efficiency gain.

\begin{table}[h]
\centering
\caption{Efficiency analysis on Lumina-mGPT-768 models, based on padding-based strategy.}
\label{tab:efficient_padding}

\footnotesize
\setlength{\tabcolsep}{3pt}
\renewcommand{\arraystretch}{1.00}
\resizebox{\textwidth}{!}{
\begin{tabular}{cccccccccc}
\bottomrule
\textbf{Models}                                   & \textbf{$\rho$}      & \textbf{Method} & \textbf{$B$} & \textbf{Batch} & \textbf{Memory} & \textbf{Mem. Saving} & \textbf{Latency} & \textbf{Through.} & \textbf{Speedup} \\ \midrule   
\multirow{3}{*}{\thead{Lumina-mGPT-768 \\($N=2352$)}} & 1                    & Full            & 2352         & 6             & 35.27GB         & -                    & 287.46s          & 0.020$\text{it/s}$ & 1.00$\times$     \\
                                                  & \multirow{2}{*}{1/8} & LineAR          &  294          &  6              &  17.31GB               & 50.92\%                     &  179.38s               &  0.033$\text{it/s}$                 & 1.65$\times$                 \\
                                                  &                      & HeadKV          &  294           &  6             &  17.59GB               & 50.12\%                     & 184.64s                 &  0.032$\text{it/s}$                 & 1.60$\times$                 \\ \toprule 
\end{tabular}
}
\vspace{-5mm}
\end{table}

\subsection{Analysis of Attention Locality and Spatial Concentration.}

To empirically validate attention locality, we analyze the distribution of attention scores over relative distances on the Lumina-mGPT-768 model. We select query tokens from 1000–1500 and examine representative layers (\emph{e.g.}, 0, 4, 10, 16, 27, and 31). To reduce noise from heterogeneous attention patterns, we focus on global heads identified by our grouping strategy.

\begin{figure}[h]
    \centering
    \includegraphics[width=1\linewidth]{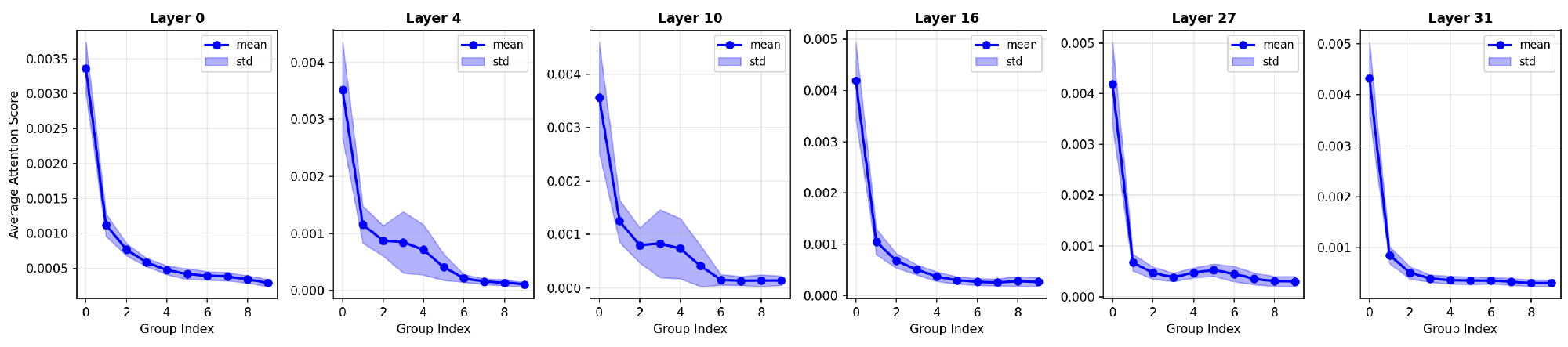}
    \caption{Attention distribution across distance bins for query positions 1000–1500.}
    \label{fig:attn_dis_across_1000_1500}
\end{figure}

For each query, we partition the token sequence into 10 equal-length bins based on relative distance to the query, and compute the mean and standard deviation of attention scores per bin. The results are shown in Fig.~\ref{fig:attn_dis_across_1000_1500}. We observe that attention mass is consistently concentrated in the first few bins (Group 0–2), which correspond to tokens close to the query. This indicates a strong locality bias in attention allocation.

Interestingly, in intermediate layers (\emph{e.g.}, layers 4 and 10), we also observe increased attention in middle-distance bins. This is consistent with the observation that mid-level representations capture broader context. Nevertheless, the attention distribution remains heavily skewed toward nearby tokens, suggesting that locality is a dominant property across layers.

\begin{figure}[h]
    \centering
    \includegraphics[width=1\linewidth]{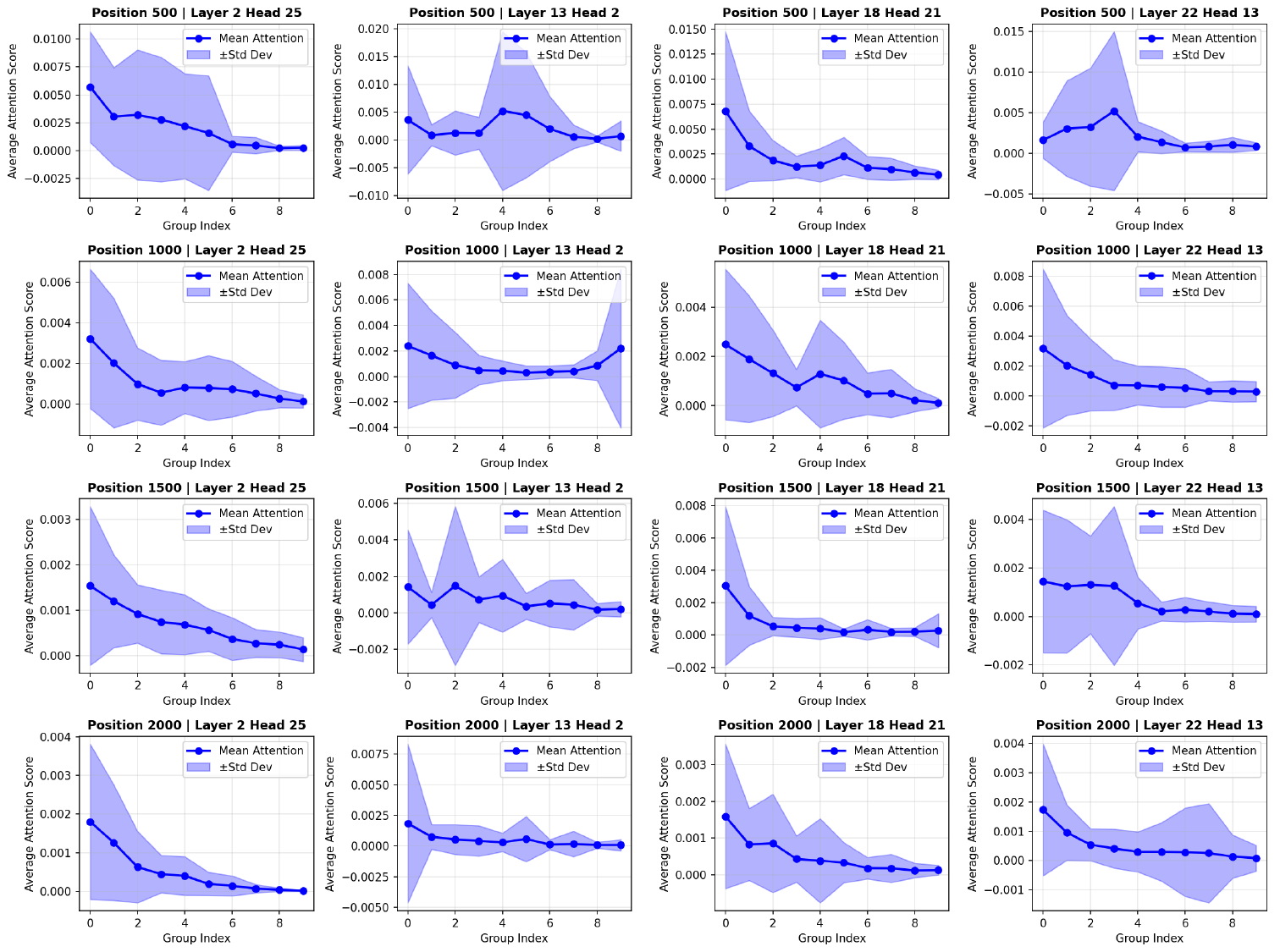}
    \caption{Attention distribution for specific layers, heads, and query positions.}
    \label{fig:att_dis_spec_layer_pos}
\end{figure}

We further conduct fine-grained analysis at the layer and head level, as shown in Fig.~\ref{fig:att_dis_spec_layer_pos}. The attention distributions for specific query positions exhibit consistent patterns, where high attention values are concentrated within a local region. Additionally, the standard deviation shows that, although distant tokens occasionally receive high attention, such cases are sparse and do not dominate the distribution.

\begin{figure}[h]
    \centering
    \includegraphics[width=1\linewidth]{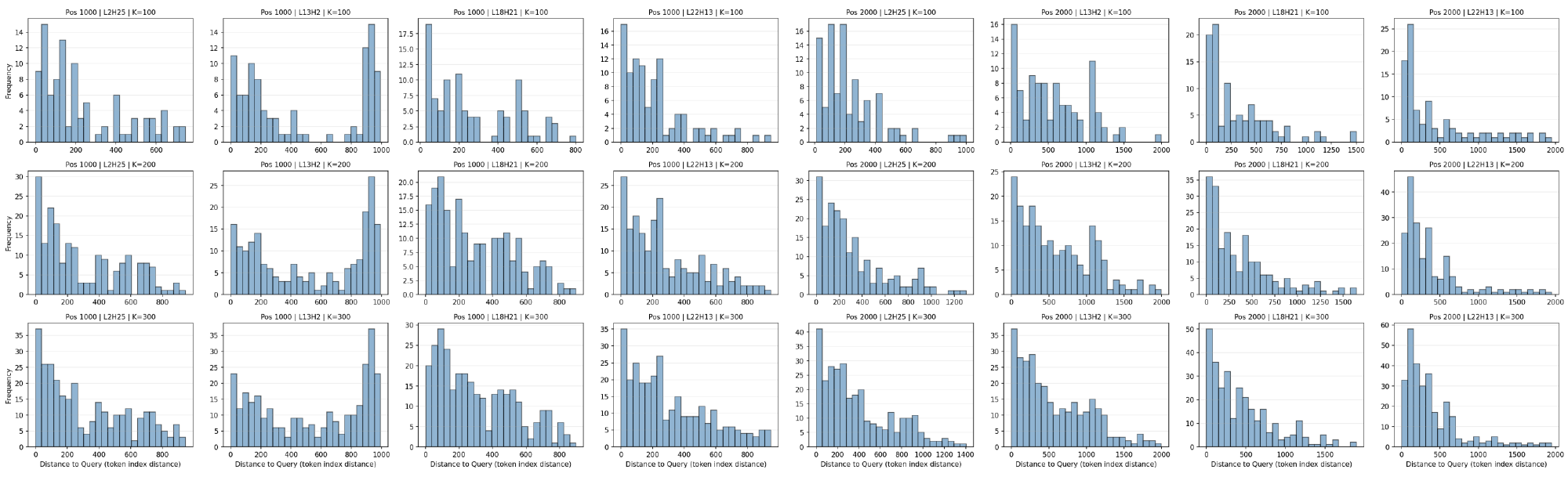}
    \caption{Spatial distribution of Top-K tokens for selected queries. }
    \label{fig:topk_distance}
\end{figure}

Finally, to examine the effect of locality bias on token selection, we visualize the distribution of Top-K tokens over relative distances for selected queries. As shown in Fig.~\ref{fig:topk_distance}, the selected tokens are predominantly located near the query, forming a concentrated local cluster. This confirms that Global Top-K strategies naturally favor locally adjacent tokens.

Overall, these observations consistently demonstrate that attention scores exhibit strong spatial concentration, which leads to a locality-biased token selection behavior.

\subsection{Theoretical Analysis: Why Stratified Token Eviction?}

To analyze the effectiveness of Stratified Token Eviction (STE), we formulate token selection as a constrained subset-optimization problem subject to a limited KV cache budget. Given a candidate token set $\Omega = \{x_1, \cdots, x_T\}$, the objective is to select a subset $S \subseteq \Omega$ with fixed budget $|S| = B$ that maximizes the retained information:
\begin{equation}
\max_{S \subseteq \Omega} I(S),
\quad \text{s.t. } |S| = B,
\end{equation}
where $I(S)$ denotes the total information retained by the selected token subset $S$.

To characterize the contribution of each token, we define the Marginal Information Gain (MIG) of adding token $x_i$ into the current subset $S$ as:
\begin{equation}
\Delta I(x_i \mid S)
=
I(S \cup \{x_i\}) - I(S).
\end{equation}

Under the optimal resource allocation principle, an ideal cache allocation strategy should prioritize tokens with the largest marginal contribution to the overall retained information. Introducing the Lagrange multiplier $\lambda$, the constrained optimization can be written as:
\begin{equation}
\mathcal{L}(S,\lambda)
=
I(S)-\lambda(|S|-B).
\end{equation}
To analyze whether adding a token $x_i$ is beneficial under the constrained budget, we consider the incremental change of the Lagrangian after including $x_i$ into the selected subset $S$:
\begin{equation}
\Delta \mathcal{L}
=
\mathcal{L}(S\cup\{x_i\},\lambda)
-
\mathcal{L}(S,\lambda).
\end{equation}

Substituting the equation yields:
\begin{equation}
\Delta \mathcal{L}
=
I(S\cup\{x_i\})-I(S)-\lambda
=
\Delta I(x_i\mid S)-\lambda.
\end{equation}

Therefore, adding token $x_i$ improves the objective only when:
\begin{equation}
\Delta I(x_i\mid S)\ge\lambda,
\label{eq:mig}
\end{equation}
which indicates that a token should be preserved only if its marginal information contribution exceeds the minimum utility threshold determined by the cache budget.

\textbf{Global Top-K Strategy.}
Existing Global Top-K strategies approximate Eq.~\ref{eq:mig} by directly using the absolute attention score $a_i$ as a surrogate for the marginal information gain:
\begin{equation}
a_i \propto \Delta I(x_i \mid S).
\end{equation}

However, this approximation becomes unreliable in visual generation due to the strong semantic redundancy among visual tokens. Existing studies~\cite{liang2026improved,so2025grouped} show that spatially adjacent visual tokens often encode highly overlapping semantic information. Consequently, attention scores tend to concentrate around a narrow local neighborhood near the query token.

From an information-theoretic perspective, the retained information function $I(S)$ can be reasonably approximated as a submodular function exhibiting diminishing-return behavior. Specifically, when semantically similar tokens are already preserved in the cache, the marginal information contribution of another nearby token decreases:
\begin{equation}
\Delta I(x_i \mid S_1)
\ge
\Delta I(x_i \mid S_2),
\quad
S_1 \subseteq S_2.
\end{equation}

As a result, although spatially adjacent tokens may still receive high attention scores due to strong query relevance, their actual marginal information gain can become insufficient under the limited cache budget:
\begin{equation}
\Delta I(x_i \mid S) < \lambda.
\end{equation}

This reveals the fundamental limitation of Global Top-K strategies: attention scores mainly capture token relevance to the query, but fail to account for redundancy with already retained tokens.

\textbf{Stratified Token Eviction Strategy.}
To alleviate this issue, our Stratified Token Eviction (STE) divides the candidate tokens into two spatial domains according to their distance to the current query:
\begin{equation}
\Omega = \Omega_n \cup \Omega_l,
\end{equation}
where $\Omega_n$ and $\Omega_l$ denote the near-range and long-range domains, respectively.

Instead of allowing all tokens to compete globally under a single cache budget, STE allocates independent budgets $k_n$ and $k_l$ to the two domains:
\begin{equation}
k_n + k_l = B.
\end{equation}

The optimization objective becomes:
\begin{equation}
\max_{S_n \subseteq \Omega_{n} \, S_l \subseteq \Omega_{l}} I_n(S_n) + I_l(S_l),
\quad
\text{s.t. }
|S_n| = k_n,
\ |S_l| = k_l,
\end{equation}
where $S_n \subseteq \Omega_n$ and $S_l \subseteq \Omega_l$ denote the selected token subsets from the near-range and long-range domains. The resulting optimality condition is:
\begin{equation}
\Delta I_n(x_i \mid S_n) \ge \lambda_n \quad \text{and} \quad \Delta I_l(x_j \mid S_l) \ge \lambda_l.
\end{equation} 
By decoupling the competition between near-range and long-range tokens, STE mitigates the domination of locally clustered high-attention tokens and prevents the cache from collapsing into a narrow spatial neighborhood. In other words, STE acts as a structural regularizer that encourages diversity-aware cache allocation.
Although spatial distance does not strictly guarantee semantic diversity, it serves as an effective inductive bias in practical visual generation scenarios, encouraging the preservation of both local detailed structures and globally distributed contextual information.

\end{document}